\documentclass{article}

\pdfoutput=1

\PassOptionsToPackage{square, numbers}{natbib}
\usepackage[final]{nips_2016}

\usepackage[utf8]{inputenc} 
\usepackage[T1]{fontenc}    
\usepackage{hyperref}       
\usepackage{url}            
\usepackage{booktabs}       
\usepackage{amsfonts}       
\usepackage{nicefrac}       
\usepackage{microtype}      
\usepackage{graphicx}
\graphicspath{{./figure/}}
\usepackage{subcaption} 
\usepackage[ruled]{algorithm2e}
\usepackage{caption}
\usepackage{cite}
\usepackage{color}
\usepackage{ctable}
\usepackage{siunitx}

\title{An Efficient Approach to Boosting\\Performance of Deep Spiking Network Training}

\author{
    Seongsik Park$^1$, Sang-gil Lee$^1$, Hyunha Nam$^1$, and Sungroh Yoon$^{1,2}$\thanks{To whom correspondence should be addressed.} \\
    $^1$Electrical and Computer Engineering,  Seoul National University,  Seoul 08826, Korea\\
    $^2$Neurology and Neuroscience, Stanford University, Stanford CA94305, USA\\
    \texttt{sryoon@snu.ac.kr} \\
}
\begin{document}
\maketitle
\begin{abstract}
Nowadays deep learning is dominating the field of machine learning with state-of-the-art performance in various application areas. Recently, spiking neural networks (SNNs) have been attracting a great deal of attention, notably owning to their power efficiency, which can potentially allow us to implement a low-power deep learning engine suitable for real-time/mobile applications. However, implementing SNN-based deep learning remains challenging, especially gradient-based training of SNNs by error backpropagation. We cannot simply propagate errors through SNNs in conventional way because of the property of SNNs that process discrete data in the form of a series. Consequently, most of the previous studies employ a workaround technique, which first trains a conventional weighted-sum deep neural network and then maps the learning weights to the SNN under training, instead of training SNN parameters directly. In order to eliminate this workaround, recently proposed is a new class of SNN named deep spiking networks (DSNs), which can be trained directly (without a mapping from conventional deep networks) by error backpropagation with stochastic gradient descent. In this paper, we show that the initialization of the membrane potential on the backward path is an important step in DSN training, through diverse experiments performed under various conditions. Furthermore, we propose a simple and efficient method that can improve DSN training by controlling the initial membrane potential on the backward path. In our experiments, adopting the proposed approach allowed us to boost the performance of DSN training in terms of converging time and accuracy.

\end{abstract}

\section{Introduction}
Recently, a breakthrough in deep learning \citep{Lecun:2015dt, Lecun:1998we} has shown that, with carefully selected learning rules and regularization methods, multiple non-linear representations from deep artificial neural networks can achieve state-of-the-art performance in various tasks including image recognition \citep{Krizhevsky:2012wl}, natural language processing \citep{2014arXiv1406.1078C, Sutskever:2014ty, 2014arXiv1409.0473B}, and bioinformatics \citep{Chicco:2014dd, lee2016deeptarget, park2016deepmirgene, Min29072016}.

While the conventional deep neural networks (DNNs) excel at the problems on many different areas, one of their major drawbacks is that they are computationally expensive: A single neuron in the layer requires all values of the neurons from the previous layer to calculate a weighted sum, which needs a huge amount of multiplicative operations. Thus, the current deep learning frameworks rely heavily on GPU-based parallel computation, making it infeasible, without some compromises, to apply them to mobile devices or (even power-efficient) chips. 

In addition, the method of backpropagation, the most popular method for training DNNs, requires a differentiable activation function in each neuron, flowing information between neurons using real numbers. However, from a neurological viewpoint, this backpropagation technique is not an efficient mechanism. Information is encoded with ``all-or-none'' action potentials, such as the ``spike train,'' and only a small subset of neurons propagate information by firing spikes. 

In this regard, spiking neural networks (SNNs) can provide a much more power-efficient way of implementing artificial neurons than conventional DNNs~ \citep{merolla2014million, Maass:1997dg} leveraged by their compact and sparse activity profiles. Triggered by the recent success of DNNs, the research on SNNs is gaining momentum with renewed interest: Various neuronal models \citep{hodgkin1952quantitative, gerstner2002spiking}, spike encoding methods \citep{gutig2006tempotron, izhikevich2006polychronization}, and learning rules \citep{kistler2000modeling, fremaux2010functional} have been proposed.

Many of existing approaches train and convert a conventional DNN to its SNN counterpart by replacing the activation functions (mostly ReLU) in the trained DNN with spiking neurons, producing the SNN. For instance, \citet{OConnor:2013et} mapped a trained deep belief network to an SNN by using the Siegert mean-firing-rate approximation model for integrate-and-fire spiking neurons. \citet{Diehl:2015ix} trained a DNN with the stochastic gradient descent (SGD) and then converted it to an SNN based on parameter optimization. \citet{Zambrano:2016vu} used the asynchronous pulsed sigma-delta coding for spike trains with adaptive dynamic range in the membrane potential. 

While these pioneering methods closed the gap of performance between conventional DNNs and SNNs, they could not train SNNs with the spikes by backpropagation with SGD. Although \citet{2016arXiv160808782H} reported that they could train SNNs with a differentiable activation function, their work incurred additional computation overhead for handling activation and gradients, thus discouraging the deployment of their method for neuromorphic hardware with limited resources.

To address the limitation of these techniques, \citet{Welling:2016vp} recently proposed deep spiking networks (DSNs), one of the first multi-layer SNNs that can be trained with spikes by backpropagation with SGD. They reported that their DSN-based deep neural network delivered performance comparable to conventional weighted-sum DNNs with similar capacity.
In addition, they proposed the method of fractional stochastic gradient descent (F-SGD), which is a training algorithm to improve SNN performance. Furthermore, most parts of their method require only additive operations rather than multiplicative ones, making their network power-efficient and hardware-friendly for implementation. The authors also made an important discovery that adjusting the initial value of the membrane potential on the backward path of an SNN could significantly affect the quality of training.

Inspired by this intriguing discovery, in this paper, we thoroughly analyze the impact of initializing the membrane potential on the overall performance of training an SNN. Based on this analysis, we propose a new, efficient approach that can boost the performance of training SNNs. Our approach is grounded on the method of precharging membrane potentials in the previous work, effectively advancing it by reducing the side effects of precharged potentials as training progresses.
Furthermore, we share our neurological insight gained by the observations of network behavior: We can interpret the performance boosts as originating from the long-term potentiation of synapses, while we can represent the negative impact of the initial membrane potentials as a trace of background activities interfering with memory maintenance in synapses.

\section{Methods and Experiments}
As baseline training algorithm, we present the original version of the training methodology for DSNs~\citep{Welling:2016vp} in Algorithm \ref{alg:pseudocode_training_dsn}.

As outlined in this algorithm, the training procedure of DSNs is different from that of conventional DNNs. For each training example, we extract a set of spike trains (series of spikes) $t$ times and use each spike train to train a DSN. In this paper, we call the quantity $t$ the \textit{\textbf{time period}} for training (in other papers, the term ``time steps'' is also frequently used). The spikes in a spike train are propagated forward and backward through the network, while the gradient information for weight updates is accumulated. After each time period is over, the parameters in the DSN are updated using the accumulated gradient information.


The properties of information encoding and propagation in SNNs make the time period for training data have a critical role. The neurons in an SNN would receive more stimuli as the time period increases since the input spike train is proportional to it. The neurons can thus have more chances to fire a spike, which would eventually improve the performance of the SNN.

\begin{algorithm}[t]
\SetAlgoLined
$E :$ \# of epochs, $D :$ \# of training data, $T :$ Time period \\
$phi_{bwd}:$ Membrane potentials on the backward path

\vspace{0.5em}

\For{e in E}{
    \For{d in D} {
        $phi_{bwd} \leftarrow \textsc{InitializationFunction} $\;
        $del \leftarrow 0$\;
        \For{t in T} {
            $s_t\leftarrow \textsc{SamplingInputData}(d)$\;
            $err_t \leftarrow \textsc{ForwardPathOperation}(s_t)$\;
            $del_t \leftarrow \textsc{BackwardPathOperation}(err_t)$\;
            
            $del \leftarrow del + del_t$\;
        }
        \textsc{UpdateParameters}$(del)$\;
    }
}
\caption{Training in Deep Spiking Networks \citep{Welling:2016vp}}
\label{alg:pseudocode_training_dsn}
\end{algorithm}

\subsection{Effects of Time Periods and Initial Potential on SNN Training}

To verify the previous work and validate the effectiveness of the proposed approach, we performed a diverse set of experiments using the code under the same default hyper-parameters and conditions from the original work (available at https://github.com/petered/spiking-mlp).

\begin{figure}[t]
\begin{subfigure}{0.5\textwidth}
    \includegraphics[width=1.0\linewidth]{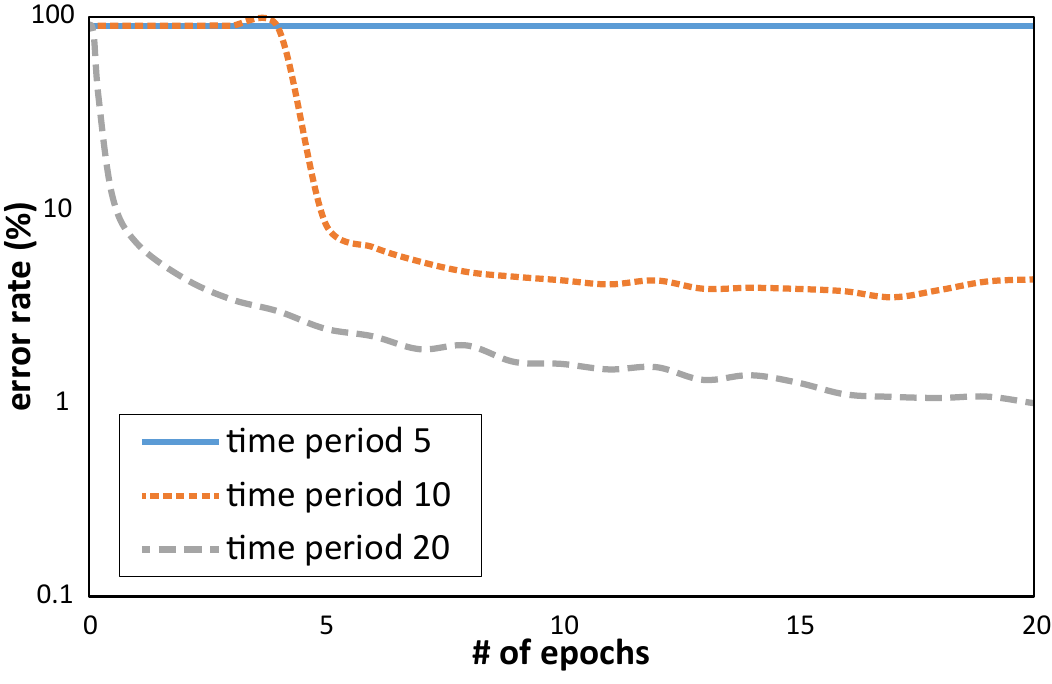}
    \caption{}
    \label{fig:training_error_time_period}
\end{subfigure}
\begin{subfigure}{0.5\textwidth}
    \includegraphics[width=1.0\linewidth]{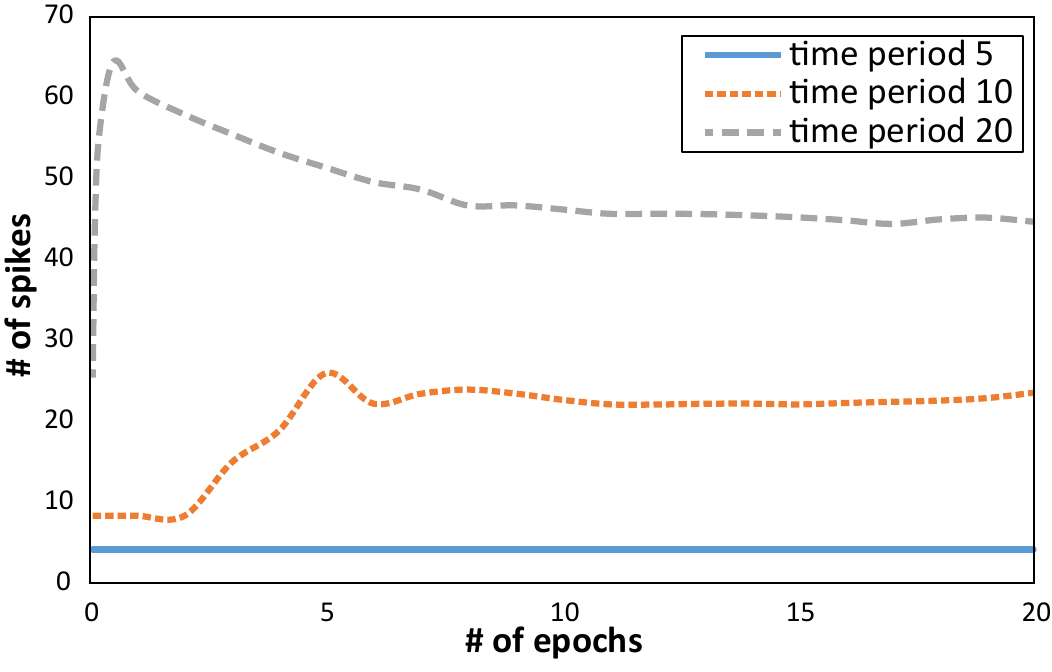}
    \caption{}
    \label{fig:num_bwd_spiking_time_period}
\end{subfigure}
\caption{(a) Training error (\%) and (b) the number of the backward spikes obtained from a DSN trained by SGD with the MNIST data using three different time periods for training (the initialization function of membrane potentials on the backward path was reset to zero).}
\end{figure}

Figure~\ref{fig:training_error_time_period} shows the training results (error rates versus epochs) obtained from training an SNN implemented as a DSN using SGD with the MNIST data~\citep{lecun1998mnist} for three different time periods. Evidently, we can observe that the time period used for training significantly affects training error rates.
%
%
As expected, the longer a time period used for training the SNN, the more accurate result we could get. In the extreme case with an (vastly insufficient) time period of 5, the SNN could not be trained at all because of the lack of spikes, as shown in Figure \ref{fig:num_bwd_spiking_time_period}. Especially, in that case, the number of backward spiking instances was insufficient to stimulate a neuron on the backward path to fire a spike. As a result, there was no change in terms of the amount of spiking regardless of the number of epochs executed. Obviously, we should use a time period that is long enough to avoid this unsatisfactory result. On the other hand, as the time period is strongly related to the total training time, it is important to make the time period as short as possible for assuring efficient training.

The main problem encountered when training an SNN with insufficient time periods is that the neurons in the SNN cannot properly propagate error signal backwards. Each neuron in an SNN integrates the spikes from other neurons and fires a spike only when its membrane potential is greater than a certain threshold. Thus, if the neuron could not obtain a sufficient number of spikes during the time period of a sampled input spike train, it will not fire. Even worse, as an SNN is getting deeper, this phenomenon of disappearing spikes between neurons would be more serious just as the vanishing gradient problem in gradient descent.
 
To resolve this problem and to improve training with relatively short time periods, the method of backwards quantization, a technique of precharging membrane potentials on the backward path, was introduced in the original DSN paper~\citep{Welling:2016vp}. The main idea is that we should precharge each membrane potential on the backward path at the beginning of each training, in order to make it easily exceed the threshold with fewer spikes. By applying this method, the spikes for error backpropagation will be fired more easily. In other words, errors will be be propagated more effectively through the backward path. Consequently, we can potentially boost the effectiveness of training in terms of convergence time and accuracy through this precharge method.

\citet{Welling:2016vp} empirically suggested three types of initialization methods: ``no reset,'' ``uniform random reset,'' and ``zero reset.'' 
%
%
In the ``no reset'' approach, the membrane potential from the previous training sample is maintained at the beginning of training using the current sample without resetting the potential. The ``uniform random reset'' and ``zero reset'' approaches refer to resetting the membrane potentials at the onset of each training with uniformly sampled random values and all zeros, respectively. Although the ``zero reset'' approach is the one that is biologically most plausible, this option makes the training of an SNN inefficient (as seen before, a DSN needs an adequate amount of time periods to deliver satisfactory performance). Consequently, the ``no reset'' or ``uniform random reset'' approaches have been commonly applied to DSN training for precharging membrane potentials on the backward path.


\subsection{Our Proposal: Step-Decay Precharge Method}

In a typical SNN, the precharged membrane potential before the beginning of training would be considered as noise, which causes faulty learning results for the current training sample. The way of precharging the membrane potential therefore critically affects the training result. 

\begin{figure}[t]
	\begin{subfigure}{0.5\textwidth}
		\includegraphics[width=1.0\linewidth]{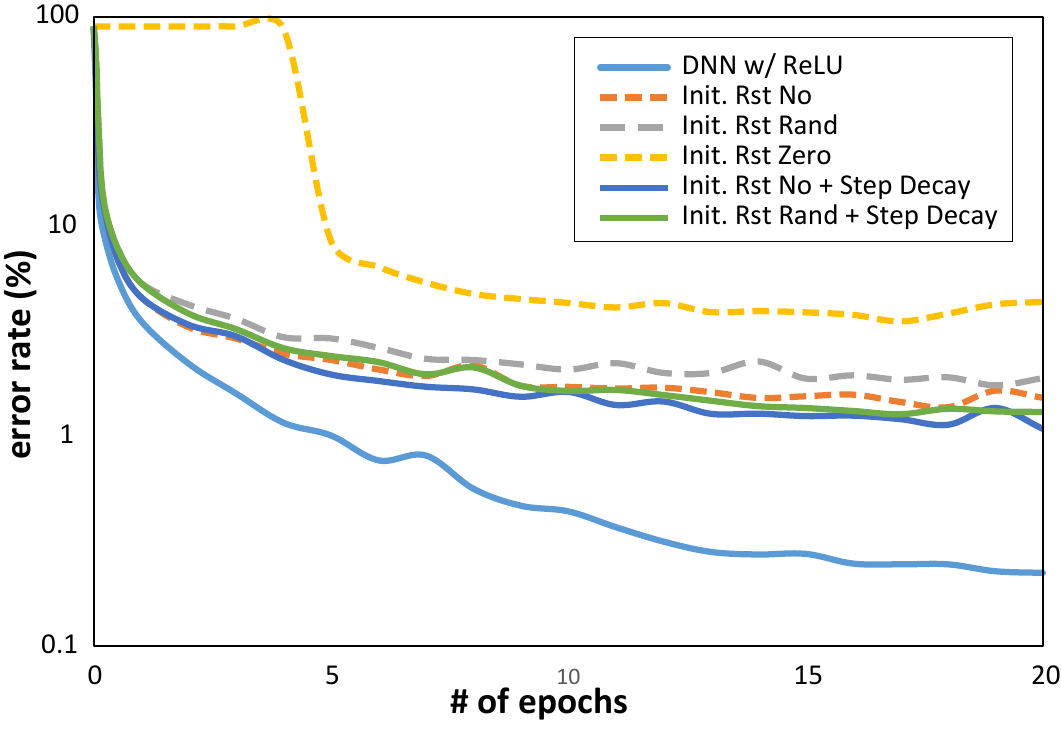}
		\caption{}
		\label{fig:training_error_sgd}
	\end{subfigure}
	\begin{subfigure}{0.5\textwidth}
		\includegraphics[width=1.0\linewidth]{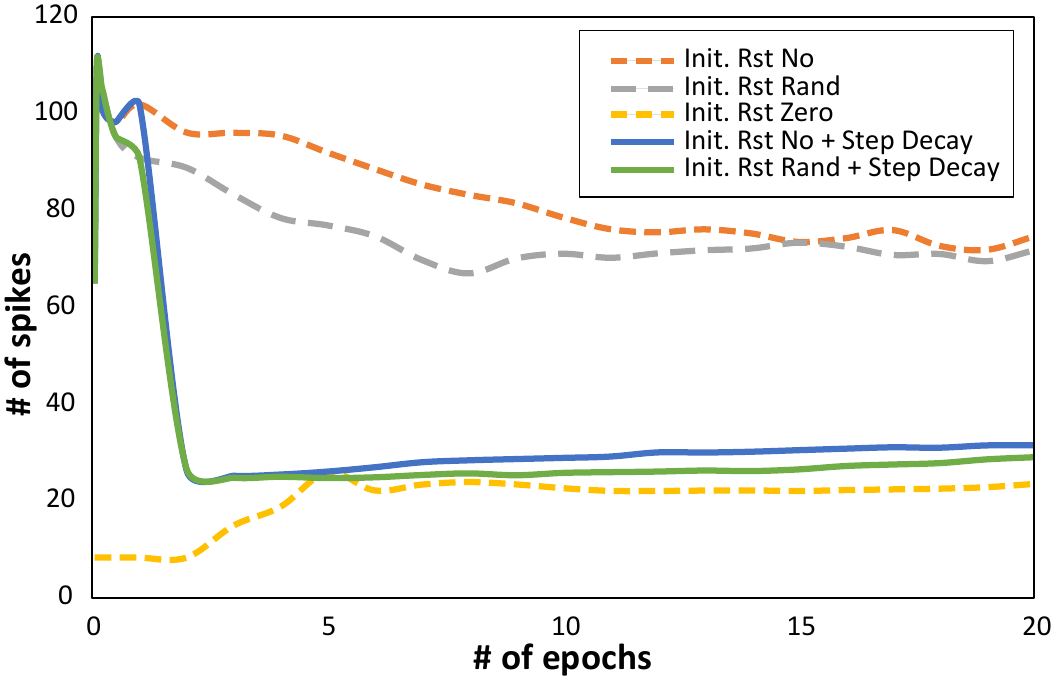}
		\caption{}
		\label{fig:num_bwd_spiking_sgd}
	\end{subfigure}
	\caption{(a) Training error (\%) and (b) the number of the backward spikes during the training of a DSN using SGD with the MNIST data (decay constant $C_d=1$ for the proposed step-decay method).}
	\label{fig:training_result_sgd}
\end{figure}

To put our proposal in a proper context, we show the SNN training results (error rates and the number of spikes versus training epochs) obtained by using each of the three aforementioned precharging methods, as shown in Figure \ref{fig:training_result_sgd}. In the figure, the curve labeled ``DNN w/ReLU'' represents the error rate of the conventional DNN with ReLU activations, and three types of dotted lines indicate the measurements obtained from using the three precharging approaches (note that two types of solid lines represent the results from using our proposed approach, which will be elaborated shortly).

Initially in the training, we noticed a similar pattern of the effects of precharged membrane potential on training quality as in the DSN paper~\citep{Welling:2016vp}. That is, the result using the ``no reset'' option gave the best result as shown in the figure. However, as the training continued, the precharged membrane potential started to become acting as noise, which hindered further training.

To address this issue, we propose a new step-wise decay method for precharging the membrane potentials. Our method can boost the effectiveness of SNN training and reduce the negative effect of precharging when an SNN completes its training. Algorithm \ref{alg:pseudocode_training_dsn_proposed} outlines our approach, in which the amount of each precharged membrane potential decreases as training epochs proceed, effectively reducing its noisy effect on the final result. As proof of concept, in this paper, we use a linear function for implementing a step-wise decay (with a linear decay constant $C_d$); other types of (linear and non-linear) functions would work as well.


\begin{algorithm}[t]
\SetAlgoLined
$E :$ \# of epochs, $D :$ \# of training data, $T :$ Time period\\$phi_{bwd} :$ Membrane potentials on the backward path, $C_d :$ Linear decay constant
\vspace{0.5em}

\For{e in E}{
    \For{d in D} {
        $phi_{bwd} \leftarrow \textsc{InitializationFunction} / (C_d \times e)$\tcp*{step-wise linear decay}
        $del \leftarrow 0$\;
        \For{t in T} {
            $s_t\leftarrow \textsc{SamplingInputData}(d)$\;
            $err_t \leftarrow \textsc{ForwardPathOperation}(s_t)$\;
            $del_t \leftarrow \textsc{BackwardPathOperation}(err_t)$\;
            
            $del \leftarrow del + del_t$\;
        }
        \textsc{UpdateParameters}$(del)$\;
    }
}
\caption{Training in Deep Spiking Networks with Proposed Step-Decay Method}
\label{alg:pseudocode_training_dsn_proposed}
\end{algorithm} 

The solid lines in Figure \ref{fig:training_error_sgd} represent the measurement results obtained by applying the proposed decay approach to SGD-based training of the same DSN shown previously ($C_d=1$). In this experiment, we could get better training results by adopting the proposed approach. For both of the ``no reset'' and ``uniform random reset'' options, our approach enhanced the training results. Compared with the original DSN, we were also able to reduce the number of backward spikes, owing to the decreased membrane potentials.



\begin{figure}[t]
	\begin{subfigure}{0.5\textwidth}
		\includegraphics[width=1.0\linewidth]{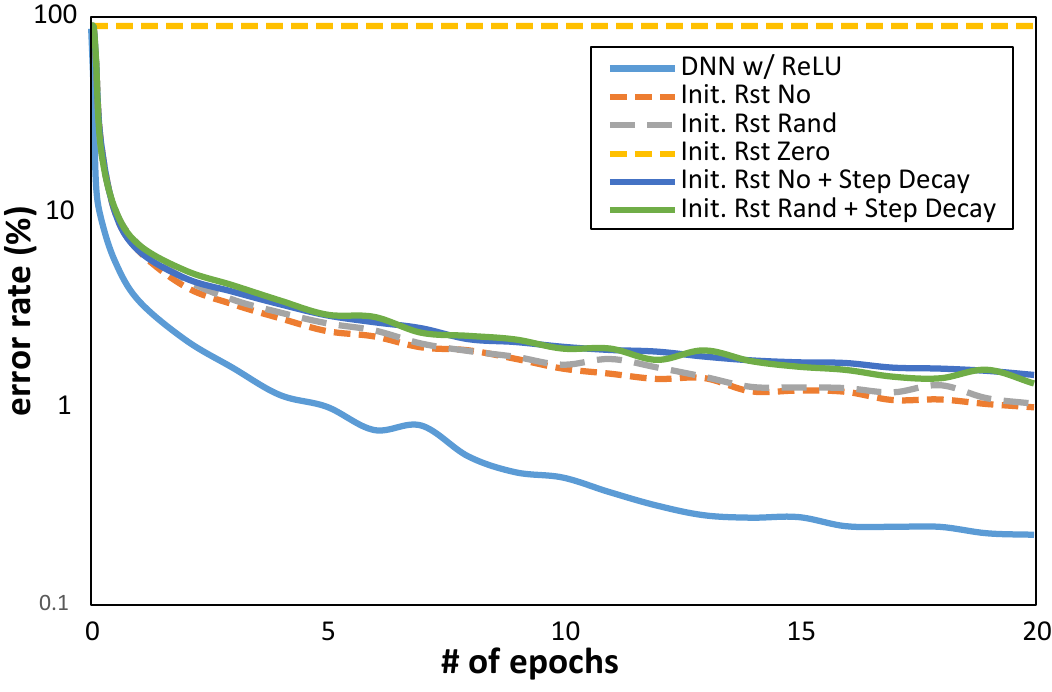}
		\caption{}
		\label{fig:training_error_fsgd}
	\end{subfigure}
	\begin{subfigure}{0.5\textwidth}
		\includegraphics[width=1.0\linewidth]{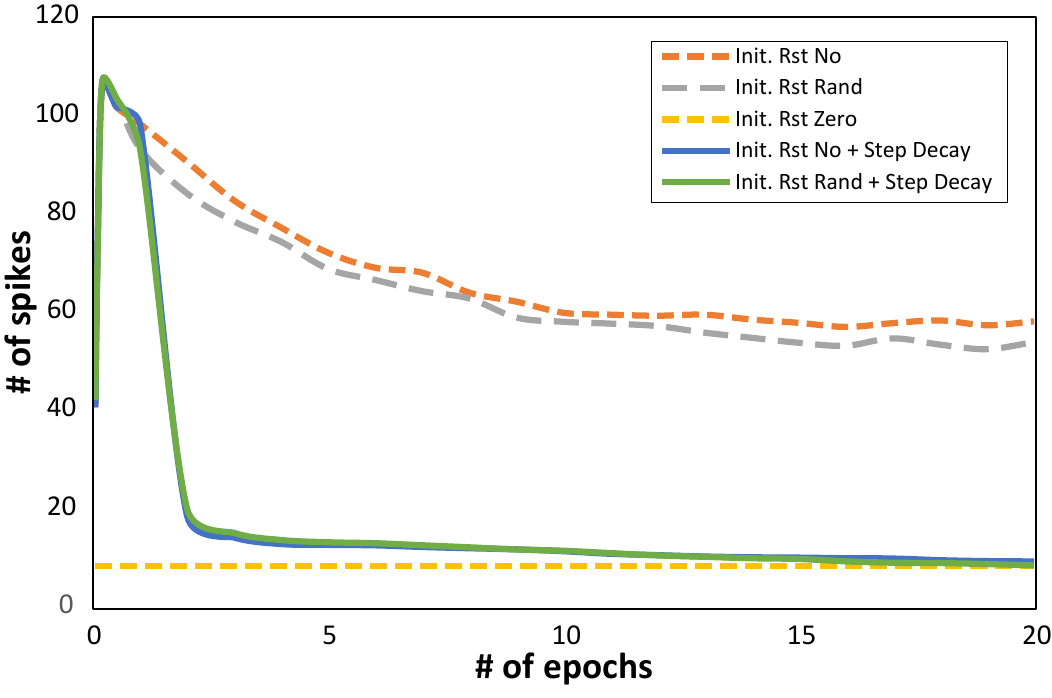}
		\caption{}
		\label{fig:num_bwd_spiking_fsgd}
	\end{subfigure}
	\caption{(a) Training error (\%) and (b) the number of the backward spikes during the training of a DSN using F-SGD with the MNIST (decay constant $C_d=1$ for the proposed step-decay method).}
\end{figure}

As shown in Figure \ref{fig:training_error_fsgd}, we further tested our approach with F-SGD, an SNN training algorithm proposed in the context DSNs, which is different from SGD in that it updates the parameters whenever a spike is generated. In contrast to the SGD case, the improvement delivered by applying our method to F-SGD was marginal. This is because the number of spikes on the backward path for F-SGD was significantly smaller than that for SGD, as seen in Figure \ref{fig:num_bwd_spiking_fsgd}. Based on this observation, we deduced that F-SGD could be more sensitive to the level of precharged membrane potentials and that we could get better training results if we reduce the effect of the decay at each epoch.

\begin{figure}[t]
\begin{subfigure}{0.5\textwidth}
    \includegraphics[width=1.0\linewidth]{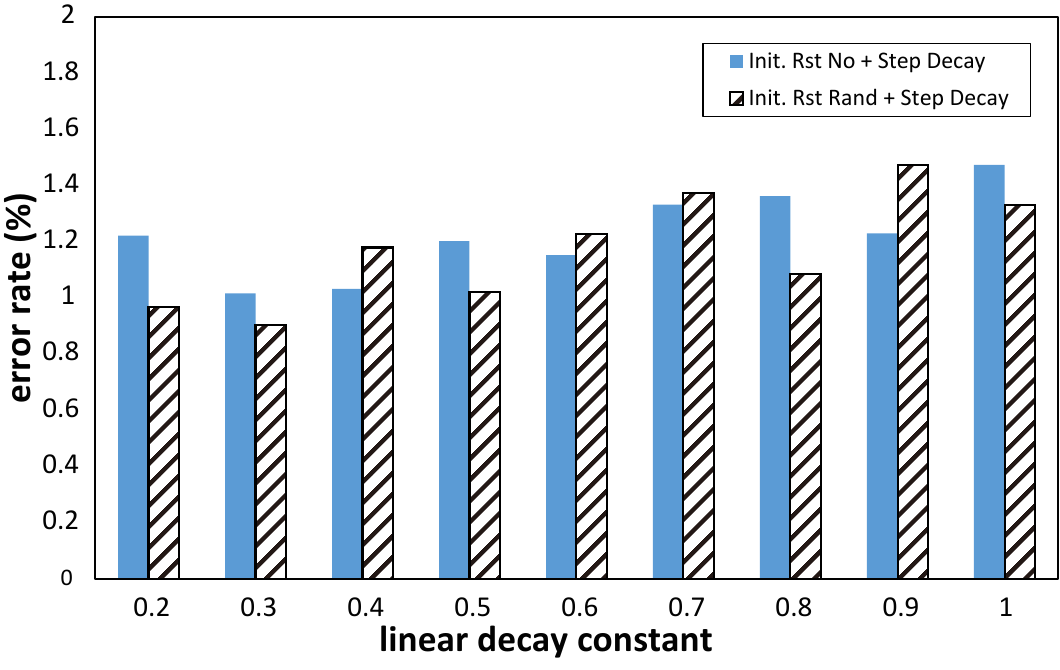}
    \caption{}
    \label{fig:training_error_fsgd_swap_linear_decay_const}
\end{subfigure}
\begin{subfigure}{0.5\textwidth}
    \includegraphics[width=1.0\linewidth]{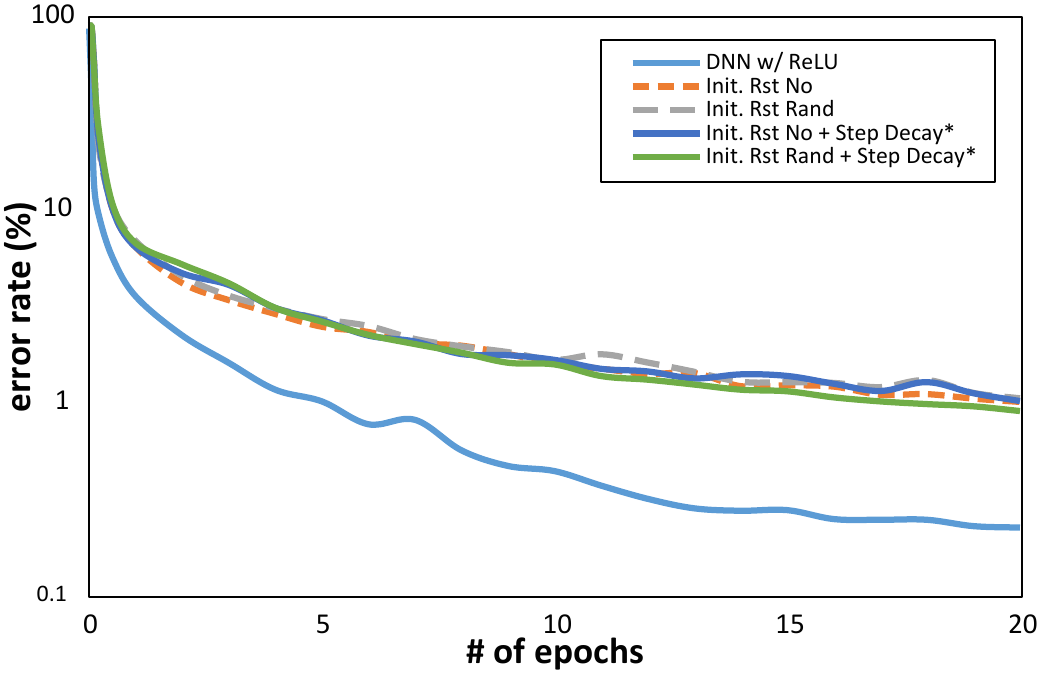}
    \caption{}
    \label{fig:training_error_fsgd_linear_decay_const} 
\end{subfigure}
\caption{(a) Training error (\%) of a DSN trained by F-SGD using the MNIST data with various values of the linear decay constant $C_d$ measured after 20 epochs (note: the case of $C_d=0.1$ is omitted because of the unsatisfactory performance obtained with this value). (b) Training error (\%) of the same DSN trained by F-SGD when $C_d=0.3$, which is empirically the best value in our experiments (data: MNIST). }
\end{figure}

Finally, to find an (empirically optimal) linear decay constant that can reduce the precharged potential effectively, we tried to sweep the values of $C_d$ from 0.1 to 0.9 (in the unit of 0.1), as shown in Figure \ref{fig:training_error_fsgd_swap_linear_decay_const}. As expected, the training accuracy varied by changing the linear decay constant, and using $C_d=0.3$ gave the best result in this specific experiment. Figure~\ref{fig:training_error_fsgd_linear_decay_const} shows the error rates versus the number of training epochs obtained by applying our approach with this best value of $C_d=0.3$ to F-SGD training of the same DSN used in the previous experiments. Table \ref{tab:Training_results_total} lists additional experimental results training obtained by using various configurations.

\ctable[
caption = {Training Results on Various Configurations with the MNIST Dataset after 20 Epochs},
label = {tab:Training_results_total}
]{lllcS[table-format=2.3]}{
	\tnote[a]{The initialization function of the membrane potentials on the backward path~\citep{Welling:2016vp}} 
	\tnote[b]{The linear decay constant (see Section 2.2)}
	
}{
	\toprule
    \multicolumn{4}{c}{\textbf{Training configurations of DSN}} & \textbf{Training error} (\%)\\		
	\midrule
	Time period & Optimizer & Initialization\tmark[a] & Step decay\tmark[b] & \\
	\midrule
	5 & SGD     &zero reset   & n/a   &90.136                                           \\
    10& SGD     &zero reset   &n/a   &4.384                                           \\
    20& SGD     &zero reset   &n/a   &1.002                                           \\ \midrule
    10& SGD     &no reset     &n/a   &1.530                                           \\
    10& SGD     &uniform random reset   &n/a   &1.898                                           \\ \midrule
    10& F-SGD   &zero reset   &n/a   &90.126                                           \\
    10& F-SGD   &no reset     &n/a   &1.006                                           \\
    10& F-SGD   &uniform random reset   &n/a   &1.050                                           \\ \midrule
    10& SGD     &no reset     &1.0    &1.088                                           \\
    10& SGD     &uniform random reset   &1.0    &1.310                                        \\ \midrule
    10& F-SGD   &no reset     &1.0    &1.474                                           \\
    10& F-SGD   &uniform random reset   &1.0    &1.332                                           \\ \midrule
    10& F-SGD   &no reset     &0.3    &1.016                                           \\
    10& F-SGD   &uniform random reset   &0.3    &0.902                                           \\
	\bottomrule
}

\section{Discussion and Conclusion}
\begin{figure}[t]
\begin{subfigure}{0.5\textwidth}
    \includegraphics[width=1.0\linewidth]{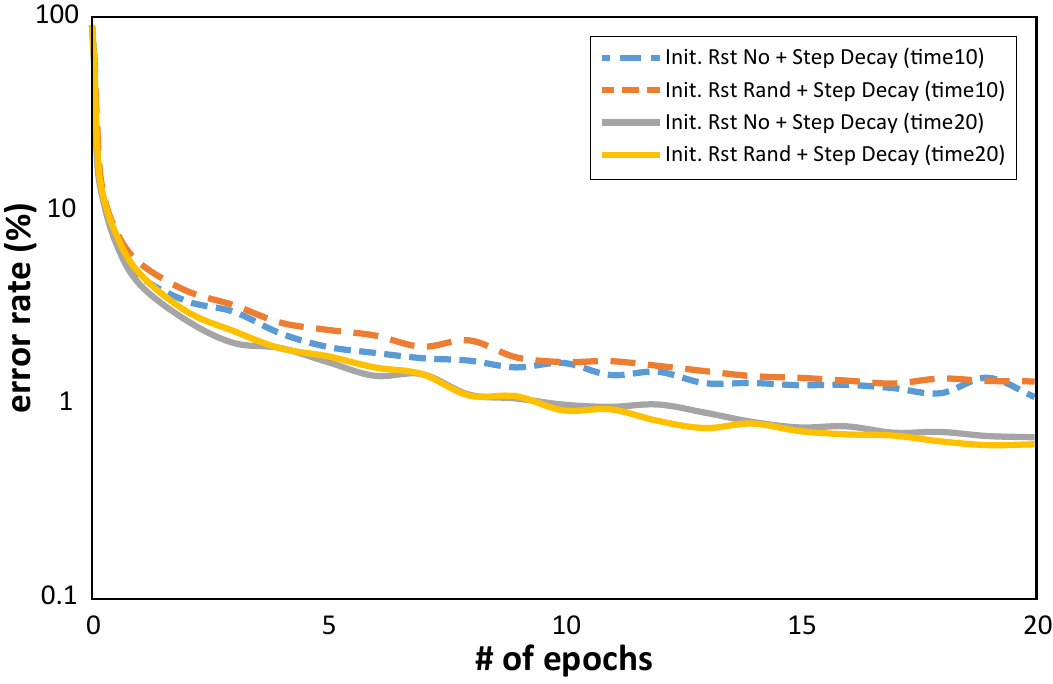}
    \caption{}
    \label{fig:training_error_sgd_step_decay_time}
\end{subfigure}
\begin{subfigure}{0.5\textwidth}
    \includegraphics[width=1.0\linewidth]{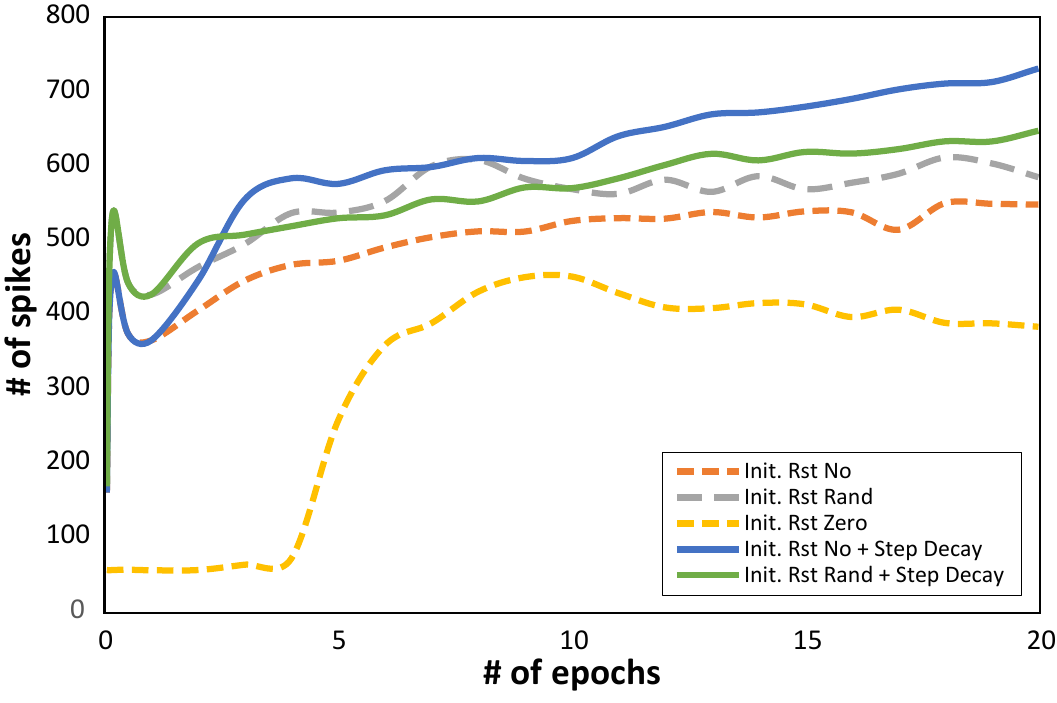}
    \caption{}
    \label{fig:num_fwd_spiking_sgd}
\end{subfigure}
\caption{(a) Training error (\%) of a DSN trained by using SGD with the MNIST data for the time periods of 10 and 20. (b) The number of forward spikes during the training of the same DSN (optimized by SGD using the MNIST data).}
\end{figure}

In this paper, we have proposed a new approach to boosting the effectiveness of DSN training with SGD and backpropagation. Our proposal is based on the fact that the initialization of membrane potentials on the backward path of a DSN significantly affects the training accuracy as previously discovered. The main idea of the proposed method is that decreasing the effect of initialization potentials can improve training accuracy as training progresses. We validated the effectiveness of our method through a diverse sets of experiments under various conditions. 

As long as we train an SNN with SGD and backpropagation (both are widely used in DNNs), the precharge method is expected to be an inevitable prerequisite for boosting the effectiveness of SNN training. In this context, we believe that our approach makes a meaningful contribution to the field, providing a simple and effective means of improving SNN training quality. 
This paper presents our experimental results only with one type of SNN (\textit{i.e.}, DSN) and one type of data (\textit{i.e.}, MNIST). We expect that our method will be able to find applications in other types of SNNs and datasets, which will be a part of our future work.


It is possible to interpret the precharge effect as a type of transfer learning, especially when we do not reset the membrane potential at the beginning of each training. In this case, the maintained membrane potential acts to compensate the lack of spiking caused by insufficient time periods. However, depending on the training data and/or the degree of network training, this could merely indicate noise, even though the maintained membrane potential led to satisfactory results in our work and in the original DSN paper. 

To scrutinize this interpretation further, we hypothesized that the randomized membrane potential with a step-wise decay could be the best combination among many initialization options. We then tried to empirically verify our hypothesis by carrying out experiments with sufficient time periods, as shown in Figure \ref{fig:training_error_sgd_step_decay_time}. 
We also found out that the firing rate of the forward path on the network increased further when we applied the step-wise decay method as depicted in Figure \ref{fig:num_fwd_spiking_sgd}. 

From a neuroscience point of view, these phenomena correspond to the long-term potentiation where a persistent strengthening of synapses occurs between neurons \citep{cooke2006plasticity}, resulting in higher spikes. In addition, if we interpret the initialized values as a trace of background activities of the network, they interfere with memory maintenance in synapses by leading a synaptic efficacy to faster decay~ \citep{higgins2014memory}, inhibiting the long-term potentiation, \textit{i.e.}, successful learning.

\section*{Acknowledgments}
This work was supported in by the BK21 Plus Project in 2016 (Electrical and Computer Engineering, Seoul National University), in part by a grant from Samsung Electronics, and in part by a grant from Naver.


\bibliographystyle{abbrvnat}
\bibliography{sample}
\end{document}